\documentclass[11pt]{article}

\usepackage{times}
\usepackage{graphicx}
\usepackage{latexsym}
\usepackage{amsfonts}
\usepackage{amssymb}
\usepackage{amsmath,bm}
\usepackage{natbib,array}
\usepackage[labelfont=bf,font=footnotesize]{caption,subfig}
\usepackage[figuresleft]{rotating}
\usepackage{multirow}
\usepackage{hyperref}

\newcommand{\xB}{\mathbf{x}}
\newcommand{\pB}{\mathbf{p}}
\newcommand{\nB}{\mathbf{n}}
\newcommand{\kB}{\mathbf{k}}
\newcommand{\XB}{\mathbf{X}}
\newcommand{\muB}{\mbox{\boldmath $\mu$}}

\newcommand{\SigmaB}{\mbox{\boldmath $\Sigma$}}
\newcommand{\OmegaB}{\mbox{\boldmath $\Omega$}}
\newcommand{\ThetaB}{\mbox{\boldmath $\Theta$}}
\newcommand{\be}{\begin{equation}}
\newcommand{\bel}[1]{\begin{equation}\label{#1}}
\newcommand{\ee}{\end{equation}}
\newcommand{\trace}{\mathrm{tr}}
\newcommand{\given}{\:|\:}
\newcommand{\ali}[1]{\begin{align}#1\end{align}}
\newcommand{\alil}[2]{\begin{align}\label{#1}#2\end{align}}
\newcommand{\SB}{\mathbf{S}}

\begin{document}

\title{Network-based clustering with mixtures of $\ell_1$-penalized Gaussian graphical models: an empirical investigation}

\author{Steven Hill\footnote{Current Address: The Netherlands Cancer Institute, Amsterdam, The Netherlands}\\
\small Centre for Complexity Science and\\[-0.8ex]
\small Department of Statistics\\[-0.8ex]
\small University of Warwick\\[-0.8ex]
\small Coventry, UK\\
\small \texttt{s.hill@nki.nl}\\
\and
Sach Mukherjee\\
\small The Netherlands Cancer Institute\\[-0.8ex]
\small Amsterdam, The Netherlands\\
\small \texttt{s.mukherjee@nki.nl}\\
}


\maketitle

\begin{abstract}
In many applications, multivariate samples may harbor previously unrecognized heterogeneity at the level of conditional independence or network structure. 
For example, in cancer biology, disease subtypes may differ with respect to subtype-specific interplay between  molecular components.
Then, both subtype discovery and estimation of subtype-specific networks present important and related challenges.
To enable such analyses, we put forward a mixture model whose components are sparse Gaussian graphical models. 
This  brings together model-based clustering and graphical modeling to permit
simultaneous estimation of cluster assignments and cluster-specific networks. 
We carry out estimation within an $\ell_1$-penalized framework, and investigate several specific penalization regimes.
We present empirical results on simulated data
and provide general recommendations for the formulation and use of mixtures of $\ell_1$-penalized Gaussian graphical models.

\vspace{0.5cm}
\end{abstract}

\newpage

\section{Introduction}

Clustering of high-dimensional data has been the focus of much statistical research over the past decade. 
The increasing prevalence of high-throughput biological data has been an important motivation for such efforts.
In molecular biology applications, clustering can be used to either group variables together \citep[e.g. to find sets of co-regulated genes;][]{Eisen1998,Toh2002}, group samples together \citep[e.g. to discover disease subtypes characterized by similar gene expression profiles;][]{Golub1999, Alizadeh2000}, or to simultaneously group both variables together and samples together \citep[`bi-clustering' methods;][]{Alon1999, Madeira2004}.
In this work we focus on the second of these approaches; that is, to cluster a small-to-moderate number of high-dimensional samples.
Numerous clustering algorithms have been used in biological applications, notably for gene expression data; see \citet{Datta2003,Thalamuthu2006, Kerr2008} and \citet{deSouto2008} for reviews and comparisons of various methods, including K-means, hierarchical clustering and model-based clustering.
Model-based clustering \citep{Fraley1998,McLachlan2002} with Gaussian mixture models is a popular approach to clustering that is rooted  in an explicit statistical model.

Another area that has received much attention in recent years is structural inference for graphical models. 
In a graphical model, a graph, comprising vertices and linking edges, is used to describe probabilistic relationships between variables; structural inference refers to estimation of the graph edge structure.
In bioinformatics applications, structural inference is  important for the elucidation of molecular networks, such as gene regulatory or protein signaling networks, from biochemical data.
Many methods for structural inference have been proposed in the literature, including those based on
Bayesian networks \citep{Friedman2000, Husmeier2003, Segal2003, Needham2007, Mukherjee2008,Hill2012c} and Gaussian graphical models \citep{Schafer2005, Toh2002, Friedman2008}.
They are reviewed, along with other approaches, in \citet{Lee2009, Hecker2009} and \citet{Markowetz2007}.

In this paper, we 
develop a model-based clustering approach with components defined by graphical models. This allows simultaneous recovery  of cluster assignments and estimation of cluster-specific graphical model structure.
Our work is of particular relevance to questions concerning undiscovered heterogeneity at the level of network structure. Such questions arise in diverse molecular biology applications.
The edge structure of biological networks can differ depending on context, e.g. disease state or other subtype, in ways that may have implications for targeted and personalized therapies \citep{Peer2011}.
When such heterogeneity is well-understood, samples can be partitioned into suitable subsets prior to network inference \citep{Altay2011} (or other supervised network-based approaches \citep{Chuang2007}).
However, in practice,
 molecular classifications that underpin such stratifications may be uncertain and moreover hitherto unknown subtypes may be present. 
In the latter case, subtype identification is itself of independent interest, as in the context of many diseases, including cancer.
A crucial observation is that if subtypes differ with respect to underlying network structure, clustering and network inference become coupled tasks. Clustering methods that do not model cluster-specific network structure (including K-means, hierarchical clustering or model-based clustering with diagonal covariance matrices \citep{deSouto2008}), 
may lead to cluster assignments that do not reflect the underlying biology and that may also compromise  the ability to elucidate network structure. 
Equally, structural inference based on the full, unclustered data can be severely confounded by the data heterogeneity. 

As cluster-specific network models, we use sparse Gaussian graphical models.
These are multivariate Gaussian models in which an undirected graph is used to represent conditional independence relationships between variables.
Inferring the edge set of a Gaussian graphical model is equivalent to identifying the location of non-zero entries in the precision matrix (see Section \ref{GGMs} below for details).
There is a rich literature on  precision matrix estimation in the context of sparse Gaussian graphical models, with the seminal paper by \citet{ Dempster1972} proposing sparse estimation by setting entries in the precision matrix to zero.
\citet{Edwards2000} provides a review of standard approaches, such as greedy stepwise backward selection, for identifying zero entries in the precision matrix.
More recent approaches have focused on using regularization, and $\ell_1$ penalization in particular, to achieve sparsity.
\citet{Meinshausen2006} use $l_1$-penalized regression \citep[lasso;][]{Tibshirani1996} to perform neighborhood selection for each node in the graph.
A sparse precision matrix can subsequently be obtained via constrained maximum likelihood estimation using the inferred sparse graph structure.
Maximum penalized likelihood estimators with an $\ell_1$ penalty applied to the precision matrix have been proposed by \cite{Yuan2007, Friedman2008, Rothman2008} and \citet{dAspremont2008}.
Analogous to the lasso, where sparse models are encouraged by shrinking some regression coefficients to be exactly zero, the $\ell_1$ penalty on the precision matrix encourages sparsity by estimating some matrix entries as exactly zero.
Since a sparse precision matrix corresponds to a sparse Gaussian graphical model structure, $\ell_1$ penalized estimation is well-suited for inference of molecular networks, where sparsity is often a valid assumption.
Moreover, regularization enables estimation in the challenging `large $p$, small $n$' regime that is ubiquitous in these settings, but renders standard covariance estimators inapplicable or ill-behaved.

Our work adds to the literature in two main ways. First, the penalized mixture-model formulation we propose extends previous  work.
\citet{Mukherjee2011} put forward a related `network clustering' approach, but 
this is not rooted in a formal statistical model and 
  estimation is carried out using a heuristic, K-means-like algorithm with `hard' cluster assignments.
  We show empirically that likelihood-based inference via an EM formulation confers benefits over this approach.
EM algorithms for penalized likelihoods have previously been proposed for finite mixture of regression models \citep{Khalili2007, Stadler2010} and for penalized model-based clustering \citep{Pan2007, Zhou2009}.
The approach in \citet{Zhou2009} is similar to the one here. However, our $\ell_1$ penalty takes a more general form, 
allowing also for dependence on mixing proportions at the level of the full likelihood.
We show that at smaller sample sizes in particular, the  $\ell_1$ penalty  we propose offers substantial gains. 
Furthermore, while we are interested in both clustering and cluster-specific network estimation, 
\citet{Zhou2009} focus  on the use of variable selection to improve clustering accuracy.

Second, we present empirical results investigating the performance of penalisation regimes. 
A penalty parameter controls the extent to which sparsity is encouraged in the precision matrix and corresponding graphical model.
The choice of method for setting the penalty parameter together with the different forms of the $\ell_1$ penalty  itself result in several possible regimes that can be difficult to choose between {\it a priori}.
Our results show that the choice of regime can be influential 
and  suggest 
general recommendations.  

The remainder of this paper is organized as follows.
In the next Section we introduce $\ell_1$-penalized estimation for Gaussian graphical models and model-based clustering, and then go on to describe the proposed mixture model.
In Section 3 we present an empirical comparison, on synthetic data, of several regimes for the $\ell_1$ penalty term and tuning parameter selection.
In Section 4 we close with a discussion of our findings and suggest areas for future work.

\section{Methods}

\subsection{Penalized estimation of Gaussian graphical model structure}\label{GGMs}
Let $\XB=(X_1,\ldots, X_p)^{\mathsf{T}}$ denote a random vector having $p$-dimensional  Gaussian density $f\!\left(\muB,\SigmaB\right)$
with mean $\muB$ and covariance matrix $\SigmaB$.
A Gaussian graphical model 
uses an undirected graph $G=(V, E)$ to describe conditional independence relationships between the random variables $X_1,\ldots, X_p$. 
The $p$ vertices $V$ of the graph are identified with $X_1,\ldots, X_p$
with edge $(i,j) \notin E$ if and only if $X_i$ is conditionally independent of $X_j$ given all other variables, or equivalently, if and only if there is zero partial correlation between $X_i$ and $X_j$ given all other variables ($\rho_{ij}=0$).
Let $\OmegaB=\SigmaB^{-1}$ denote the inverse covariance or precision matrix, and let $\omega_{ij}$ be entry $(i,j)$ of $\OmegaB$.
Then, the relationship between $\OmegaB$ and partial correlations is given by $\rho_{ij}=-\frac{\omega_{ij}}{\sqrt{\omega_{ii}\omega_{jj}}}$.
Therefore, non-zero entries in $\OmegaB$ correspond to edges in the GGM, that is $\omega_{ij}\neq0 \iff (i,j) \in E$.
 Thus, inferring the edge set of a GGM is equivalent to identifying the location of non-zero entries in the precision matrix.
 
Suppose $\xB_1,\ldots,\xB_n$, with $\xB_i=\left(x_{i1},\ldots,x_{ip}\right)^{\mathsf{T}}$ is a random sample from $f\!\left(\muB,\SigmaB\right)$.
Let $\bar{\xB}=\frac{1}{n}\sum_{i=1}^n{\xB_i}$ denote sample mean
and $\hat{\SigmaB}=\frac{1}{n}\sum_{i=1}^{n}{(\xB_i-\bar{\xB})(\xB_i-\bar{\xB})^{\mathsf{T}}}$  sample covariance. 
The precision matrix $\OmegaB$ may be estimated by maximum likelihood. The log-likelihood function is given, up to a constant,  by
\be
l(\OmegaB) = \log \left|\OmegaB\right| - \trace(\OmegaB \hat{\SigmaB})
\ee
where $\left|\cdot\right|$ and $\trace(\cdot)$ denote matrix determinant and trace respectively.
The maximum likelihood estimate is given by inverting the sample covariance matrix, $\hat{\OmegaB}=\hat{\SigmaB}^{-1}$.
However for $n<p$, $\hat{\SigmaB}$ is singular and so cannot be used to estimate $\OmegaB$.
Even when $n\geq p$, $\hat{\OmegaB}$ can be a poor estimator for large $p$ and does not in general yield sparse precision matrices.

Sparse estimates can be encouraged by placing an $\ell_1$ penalty on the entries of the precision matrix $\OmegaB$.
This results in the following penalized log-likelihood:
\bel{pLL1}
l_p(\OmegaB) = \log \left|\OmegaB\right| - \trace(\OmegaB \hat{\SigmaB}) - \lambda\left\|\OmegaB\right\|_1
\ee
where $\left\|\OmegaB\right\|_1=\sum_{i,j}{\left|\omega_{i,j}\right|}$ is the elementwise $\ell_1$ matrix norm and $\lambda$ is a non-negative tuning parameter controlling sparsity of the estimate.
The maximum penalized likelihood estimate is obtained by maximizing (\ref{pLL1}) over symmetric, positive-definite matrices.
This is a convex optimization problem and several procedures have been proposed to obtain solutions.
\citet{Yuan2007} used the maxdet algorithm, while 
\citet{dAspremont2008} proposed a more efficient semi-definite programming algorithm using interior point optimization.
\citet{Rothman2008} offered a fast approach employing Cholesky decomposition and the local quadratic approximation, and
\citet{Friedman2008} proposed the even faster graphical lasso algorithm, based on the coordinate descent algorithm for the lasso.
We use the graphical lasso algorithm in our investigations and refer the interested reader to the references for full details.

\subsection{Gaussian mixture models}

We now suppose $\xB_1,\ldots,\xB_n$ is a random sample from a finite Gaussian mixture distribution,
\be
f(\xB_i;\ThetaB) = \sum_{k=1}^K{\pi_k f_k(\xB_i \given \muB_k, \SigmaB_k)}
\ee
where the mixing proportions $\pi_k$ satisfy $0\leq \pi_k \leq 1$ and $\sum_{k=1}^K{\pi_k}=1$, 
$f_k$ is the $p$-dimensional multivariate Gaussian density with component-specific mean $\muB_k$ and covariance $\SigmaB_k$, 
and $\ThetaB=\left\{(\pi_k, \muB_k, \SigmaB_k) : k=1,\ldots,K\right\}$ is the set of all unknown parameters.
The log-likelihood for the sample is given by
\bel{unpen}
l(\ThetaB) = \sum_{i=1}^n{\log \left(\sum_{k=1}^K{\pi_k f_k(\xB_i \given \muB_k, \SigmaB_k)}\right)}.
\ee
Maximizing this log-likelihood is difficult due to its non-convexity.
The Expectation-Maximization (EM) algorithm \citep{Dempster1977} can be used to obtain maximum likelihood estimates.

In model-based clustering \citep{Fraley1998, McLachlan2002}, each mixture component corresponds to a cluster. 
In the present setting, since each cluster (or component) is Gaussian distributed with a cluster-specific (unconstrained) 
covariance matrix, each cluster represents a distinct Gaussian graphical model.

\subsection{Mixture of penalized Gaussian graphical models}

In the Gaussian mixture model with cluster-specific covariance matrices, the number of parameters is of order $Kp^2$.
Estimation is more challenging than for a single precision matrix (or Gaussian graphical model) and so,
as described above, in settings where number of variables $p$ is moderate-to-large in relation to sample size $n$, overfitting and invalid covariance estimates are a concern.
We employ an $\ell_1$ penalty on each of the $K$ precision matrices to promote sparsity and ameliorate these issues.
Such $\ell_1$ penalties have previously been proposed for clustering with Gaussian graphical models \citep{Zhou2009, Mukherjee2011}.

We propose the following penalized log-likelihood,
\bel{pen}
l_p(\ThetaB) = \sum_{i=1}^n {\log \left( \sum_{k=1}^K \pi_k f_k\left( \xB_i \given \muB_k, \SigmaB_k \right) \right)} - \frac{n}{2}p_{\lambda,\gamma}(\ThetaB) 
\ee
where the penalty term is given by

\bel{penterm}
p_{\lambda,\gamma}(\ThetaB) =  \lambda\sum_{k=1}^K { \pi_k^{\gamma}\left\| \OmegaB_k \right\|_1}
\ee 
and $\gamma$ is a binary parameter controlling the form of the penalty term.
Setting $\gamma=0$ results in the conventional penalty term, as used in \citet{Zhou2009}, with no dependence on the mixing proportions $\pi_k$.
Setting $\gamma=1$ weights the penalty from each cluster by its corresponding mixing proportion.
While this form of penalty is novel in this setting, an analogous penalty  has been proposed by \citet{Khalili2007} and \citet{Stadler2010} for $\ell_1$-penalized finite mixture of regression models.
In this work, we empirically compare these two forms of penalty term for clustering with, and estimation of, Gaussian graphical models.

\subsection{Maximum penalized likelihood}
As with the unpenalized log-likelihood (\ref{unpen}), the penalized likelihood (\ref{pen}) can be maximized using an EM algorithm, which we now describe.
Our algorithm is similar to that of \citet{Zhou2009}, but they consider only the $\gamma=0$ regime and also penalize the mean vectors to perform variable selection.

Let $z_{i}$ be a latent variable satisfying $z_{i}=k$ if observation $\xB_i$ belongs to cluster $k$.
Then we have $P(z_i=k)=\pi_k$ and $p(\xB_i \given z_i=k )=f_k(\xB_i \given \muB_k, \SigmaB_k)$.
The penalized log-likelihood for the complete data $\left\{\xB_i, z_i\right\}_{i=1}^n$ is
\bel{penc}
l_{p,c}(\ThetaB) = \sum_{i=1}^n { {\log(\pi_{z_i}) + \log \left(f_{z_i}\left( \xB_i \given \muB_{z_i}, \SigmaB_{z_i} \right)\right)}} - \frac{n}{2}p_{\lambda,\gamma}(\ThetaB) .
\ee

In the E-step of the EM, given current estimates of the parameters $\ThetaB^{\left(t\right)}$, we compute
\ali{
Q(\ThetaB \given \ThetaB^{\left(t\right)}) &= \mathbb{E}\left[l_{p,c}(\ThetaB) \given \left\{\xB_i\right\}_{i=1}^n, \ThetaB^{\left(t\right)}\right] \nonumber\\
                                           &= \sum_{i=1}^n {\sum_{k=1}^K {\tau_{ik}^{\left(t\right)} \left[\log(\pi_{k}) + \log \left(f_{k}\left( \xB_i \given \muB_{k}, \SigmaB_{k} \right)\right)\right]}} - \frac{n}{2}p_{\lambda,\gamma}(\ThetaB) 
}
where $\tau_{ik}^{\left(t\right)}$ is the posterior probability of observation $\xB_i$ belonging to cluster $k$,
\bel{resps}
\tau_{ik}^{\left(t\right)} = \frac{\pi_k^{\left(t\right)} f_k\left( \xB_i \given \muB_{k}^{\left(t\right)}, \SigmaB_{k}^{\left(t\right)} \right)}{\sum_{j=1}^K {}\pi_j^{\left(t\right)} f_j\left( \xB_i \given \muB_{j}^{\left(t\right)}, \SigmaB_{j}^{\left(t\right)} \right)}
\ee
and can be thought of as a `soft' cluster assignment.

In the M-step we seek to maximize $Q(\ThetaB \given \ThetaB^{\left(t\right)})$ with respect to $\ThetaB$ to give new estimates for the parameters $\ThetaB^{\left(t+1\right)}$.
When $\gamma=0$ the mixture proportions $\pi_k$ do not appear in the penalty term $p_{\lambda,\gamma}(\ThetaB)$ and so we use the following standard EM update for unpenalized Gaussian mixture models:
\bel{pik}
\pi_k^{\left(t+1\right)} = \frac{\sum_{i=1}^n \tau_{ik}^{(t)}}{n}.
\ee
For $\gamma=1$, since $\pi_k$ appears in the penalty term, maximization of $Q(\ThetaB \given \ThetaB^{\left(t\right)})$ with respect to $\pi_k$ is non-trivial.
We follow \citet{Khalili2007} and use the standard update (\ref{pik}).
If the standard update improves $Q(\ThetaB \given \ThetaB^{\left(t\right)})$ then this is sufficient to obtain (local) maxima of (\ref{pen}).
An improvement is not guaranteed here, but as found in \citet{Khalili2007}, the method works well in practice.

Since the penalty term is independent of $\muB_k$, we again use the standard update,
\be
\muB_k^{\left(t+1\right)} = \frac{\sum_{i=1}^n{\tau_{ik}^{\left(t\right)}\xB_i}}{\sum_{i=1}^n{\tau_{ik}^{\left(t\right)}}}.
\ee

The update for $\SigmaB_k$, or equivalently $\OmegaB_k$, is given by
\alil{ICk}{
\OmegaB_k^{(t+1)} & = \operatorname*{arg\,max}_{\OmegaB_k}\ \left[\sum_{i=1}^n \tau_{ik}^{(t)} \left( \log\left| \OmegaB_k \right| - 
\trace(\OmegaB_k \SB_k^{\left(t\right)}) \right)
-n\lambda\left(\pi_k^{(t+1)}\right)^{\gamma}\left\| \OmegaB_k \right\|_1 \right] \nonumber\\
& =  \operatorname*{arg\,max}_{\OmegaB_k}\ \left[ \log\left| \OmegaB_k \right| - 
\trace(\OmegaB_k \SB_k^{\left(t\right)})
-\tilde{\lambda}_{k}^{(t)}\left\| \OmegaB_k \right\|_1\right]
}
where
\bel{ICk2}
\SB_k^{\left(t\right)} = 
\frac{\sum_{i=1}^n \tau_{ik}^{(t)}\left(\xB_i-\muB_k^{(t+1)}\right)\left(\xB_i-\muB_k^{(t+1)}\right)^{\mathsf{T}}}{\sum_{i=1}^n \tau_{ik}^{(t)}}
\ee
is the standard EM update for $\SigmaB$ and
\be
\tilde{\lambda}_{k}^{(t)}= n\lambda\frac{\left(\pi_k^{(t+1)}\right)^{\gamma}}{\sum_{i=1}^n \tau_{ik}^{(t)}}.
\ee
The optimization problem in (\ref{ICk}) is of the form of that in (\ref{pLL1}) with $\hat{\SigmaB}$ replaced by $\SB_k^{\left(t\right)}$ and a scaled tuning parameter $\tilde{\lambda}_{k}^{(t)}$.
Hence we can use the efficient graphical lasso algorithm \citep{Friedman2008} to perform the optimization.

From (\ref{pik}) we have
\bel{EMparam}
\tilde{\lambda}_k^{(t)} = \left\{ \begin{array}{ll}
\frac{\lambda}{\pi_k^{(t+1)}} & \textrm{if $\gamma=0$}\\
\lambda                     & \textrm{if $\gamma=1$}
\end{array} \right.
\ee
Hence, when $\gamma=0$, $\tilde{\lambda}_k^{(t)}$ is a cluster-specific parameter inversely proportional to effective cluster sample size, whereas $\gamma=1$ simply yields $\lambda$.
We note that, even though $\gamma=1$ gives a cluster-specific tuning parameter in the penalized log-likelihood (\ref{pen}) while $\gamma=0$ does not, the converse is actually true for the EM updates (\ref{ICk}).

Our overall algorithm is as follows:

\begin{enumerate}
	\item  Initialize $\ThetaB^{(0)}$: Randomly assign each observation $\xB_i$ into one of $K$ clusters, subject to a minimum cluster size $n_{\min}$. Set $\pi_k^{(0)}=n_k/n$ where $n_k$ is the number of observations assigned to cluster $k$, set $\muB_k^{(0)}$ to sample mean of cluster $k$, and set $\OmegaB_k^{(0)}$ to the maximum penalized likelihood estimate for the cluster $k$ precision matrix (using (\ref{pLL1})).
	\item E-step: Calculate posterior probabilities (`soft' assignments) $\tau_{ik}^{(t)}$ using (\ref{resps}).
	\item M-step: Calculate updated parameter estimates $\ThetaB^{(t+1)}$ using (\ref{pik})-(\ref{EMparam}).
	\item Iterate or terminate: Increment $t$. Repeat steps 2 and 3, or stop if one of the following criteria is satisfied:
	
\begin{itemize}
	\item A maximum number of iterations $T$ is reached; $t>T$.
	\item A minimum cluster size $n_{\min}$ is reached; $\sum_{i=1}^n \tau_{ik}^{(t)}<n_{\min}$ for some $k$.
	\item Relative change in penalized log-likelihood is below a threshold $\epsilon$; \\
	$\left|l_p(\ThetaB)^{(t)}/l_p(\ThetaB)^{(t-1)}-1\right|\leq\epsilon$.
\end{itemize}

\end{enumerate}
In all experiments below we set $T=100$, $n_{\min}=4$ and $\epsilon=10^{-4}$.
Since the EM algorithm may only find local maxima, we perform 25 random restarts and select the one 
giving the highest penalized log-likelihood.
`Hard' cluster assignments are obtained by assigning observations to the cluster $k$ with largest  probability $\tau_{ik}$.

\subsection{Tuning parameter selection}
Two approaches are commonly used to set the tuning parameter: cross-validation (CV) and criteria such as BIC.
In multifold CV, the data samples are partitioned into $M$ data subsets, denoted by $\XB^{(m)}$ for $m=1,\ldots,M$.
Let $\hat{\ThetaB}_{\lambda}^{(-m)}=\left\{(\pi_{k\lambda}, \muB_{k\lambda}, \SigmaB_{k\lambda}) : k=1,\ldots,K\right\}$ denote the penalized likelihood estimate, obtained using tuning parameter $\lambda$ and by application of the EM algorithm described above to all data save that in subset $\XB^{(m)}$ (training data).
Performance of this estimate is assessed using the predictive log-likelihood; that is, Equation (\ref{unpen}) applied to subset $\XB^{(m)}$  (test data).
This is repeated $M$ times, allowing each subset to play the role of test data.
The CV score is 
\bel{CV}
\mathrm{CV}(\lambda) = \sum_{m=1}^M\  {\sum_{i : \xB_i\in \XB^{(m)}} {\log \left(\sum_{k=1}^K {\hat{\pi}_{k\lambda}^{(-m)} f_k\left(\xB_i \given \hat{\muB}_{k\lambda}^{(-m)}, \hat{\SigmaB}_{k\lambda}^{(-m)}\right)}\right)}}.
\ee
Then we choose $\lambda$ that maximizes CV$(\lambda)$, where the maximization is performed via a grid search.
Finally, the selected value is used to learn penalized likelihood estimates from all data.

In the larger sample case, an alternative to multifold CV is to partition the data into two and perform a single train/test iteration, selecting $\lambda$ that maximizes the predictive log-likelihood on the test data with penalized parameter estimates from the training data.

We define the following BIC score for our penalized mixture model:
\bel{BIC}
\mathrm{BIC}(\lambda) = -2 l(\hat{\ThetaB}_{\lambda}) + \mathrm{df}_{\lambda}\log(n)
\ee
where $l(\cdot)$ is the unpenalized log-likelihood (\ref{unpen}), $\hat{\ThetaB}_{\lambda}$ is the penalized likelihood estimate obtained with tuning parameter $\lambda$ and $\mathrm{df}_{\lambda}$ is degrees of freedom. 
\citet{Yuan2007} proposed an estimate of the degrees of freedom for $\ell_1$-penalized precision matrix estimation, which generalizes to our penalized Gaussian mixture model setting to give
\be
\mathrm{df}_{\lambda} = K(p+1)-1 + \sum_{k=1}^K{\#\left\{(j,j') : j\leq j', \left(\hat{\omega}_{k\lambda}\right)_{jj'}\neq 0 \right\}}.
\ee
where $\left(\hat{\omega}_{k\lambda}\right)_{jj'}$ is element $(j,j')$ in $\hat{\OmegaB}_{k\lambda}$, the penalized likelihood estimate for the cluster $k$ precision matrix, using tuning parameter $\lambda$.
Using a grid search, we choose $\lambda$ that minimizes BIC($\lambda$).

BIC is often preferred over CV as it is  less computationally intensive.
However, we note that, even BIC can be computationally expensive when used within clustering since 
each $\lambda$ value in the grid search requires a full application of EM-based clustering.
Hence, to reduce computation time, we also consider a heuristic, approximate version of these approaches.
The heuristic we propose relies on the notion that the optimal tuning parameter value does not depend
strongly on cluster assignments but rather largely on general properties of the data (such as $p$ and $n$).
The approach proceeds as follows. 
First, 
observations are randomly assigned to clusters, producing $K$ pseudo-clusters each with mean size $n/K$.
Second, parameter estimates are obtained for the pseudo-clusters.
$\hat{\pi}_k$ is taken to be the proportion of samples in pseudo-cluster $k$ and $\hat{\muB}_k$ is the sample mean of pseudo-cluster $k$.
Then, for varying $\lambda$, we obtain penalized estimates $\hat{\OmegaB}_{k\lambda}$ by optimizing (\ref{pLL1}) for each pseudo-cluster with the graphical lasso.
This can be done efficiently using the \texttt{glassopath} algorithm in R \citep{Friedman2008} which obtains penalized estimates for all considered values of $\lambda$ simultaneously.
Third, using these estimates, CV (BIC) scores are calculated and maximized (minimized) to select $\lambda$.
These three steps are repeated multiple times and $\lambda$ values obtained are averaged to produce a final value.

\section{Simulated data}

In this section we apply the $\ell_1$-penalized Gaussian graphical model clustering approach to simulated data.
We consider a number of combinations of $\ell_1$ penalty term and tuning parameter scheme (as described in Methods above) and assess their performance in carrying out three related tasks.
First, recovery of correct cluster assignments. 
Second, estimation of cluster-specific graphical model structure (i.e. location of non-zero entries in cluster-specific precision matrices). Third, estimation of cluster-specific precision matrices (i.e. estimation of matrix elements, not just  locations of non-zero entries).
We note that this latter task is of less interest here since we are mainly concerned with clustering and inference of cluster-specific network structure.

\subsection{Data generation}
In our simulation we considered $p$-dimensional data consisting of $K=2$ clusters, each with a known and distinct Gaussian graphical model structure (i.e. sparse precision matrix).
Sparse precision matrices were created using an approach based on that used by \citet{Rothman2008} and \citet{Cai2011}.
In particular, we created a symmetric $p \times p$ matrix $B_1$ with zeros everywhere except for $p$ randomly chosen pairs of symmetric, off-diagonal entries, which took value 0.5.
A second matrix $B_2$ was created from $B_1$ by selecting half of the $p$ non-zero symmetric pairs at random and relocating them to new randomly chosen symmetric positions.
We then set $\OmegaB_k = B_k + \delta_k I$, where $\delta_k$ is the minimal value such that $\OmegaB_k$ is positive-definite with condition number less than $p$.
Finally, the precision matrices $\OmegaB_k$ were standardized to have unit diagonals.
This resulted in cluster-specific Gaussian graphical models each with $p$ edges, half of which were shared by both network structures.
Data were generated from $\mathcal{N}\left(\textbf{0},\OmegaB_1^{-1}\right)$ and $\mathcal{N}\left(\frac{\alpha}{\sqrt{p}}\textbf{1},\OmegaB_2^{-1}\right)$ for clusters 1 and 2 respectively, where $\mathbf{1}$ is the vector of ones.
The mean of cluster two is defined such that the parameter $\alpha$ sets the Euclidean distance between the cluster means.
In the experiments below we consider $p=25,50,100$ and cluster sample sizes of $n_k=15,25,50,100,200$.
We set $\alpha=3.5$, resulting in individual component-wise means for cluster two of 0.70, 0.50 and 0.35 for $p=25,50$ and 100 respectively.
This reflects the challenging scenario where clusters do not have substantial differences in mean values, but display heterogeneity in network structure while also sharing some network structure across clusters.

\subsection{Methods and regimes}
We assessed ability 
to recover correct cluster assignments from 50 simulated datasets, under the following four regimes for the penalty term $p_{\lambda,\gamma}(\ThetaB)$ in (\ref{penterm}): $\gamma=0$ or 1 and $\lambda$ set by BIC or a train/test scheme, maximizing the predictive log-likelihood on an independent test dataset with cluster sample sizes matching the training dataset.
These regimes are described fully above and summarized in Table~\ref{Tab:methods}.
We also compared with (i) K-means; 
(ii) standard non-penalized full-covariance Gaussian mixture models estimated using EM; and
(iii) `network clustering', an $\ell_1$-penalized Gaussian graphical model clustering approach proposed by \citet{Mukherjee2011}. 
This is 
 similar  to the approach employed here but uses a heuristic, K-means-like algorithm with `hard' cluster assignments rather than a mixture-model formulation with EM.
For (i) we used the \texttt{kmeans} function in the MATLAB statistics toolbox with K=2 and 1000 random initializations and for (iii) we used MATLAB function \texttt{network\_clustering} \citep{Mukherjee2011}.
For (ii) and (iii) we used the same stopping criteria as described in Methods above (namely, $T=100$, $n_{\min}=4$ and $\epsilon=10^{-4}$) and again carried out 25 random restarts.
Method (iii) requires maximization of $K$ penalized log-likelihoods of form (\ref{pLL1}) above (one for each cluster). For setting penalty parameters for this method, we considered either a single tuning parameter $\lambda$ shared across both clusters and set by BIC or train/test, or
cluster-specific tuning parameters $\lambda_k$, set analytically before each call to the penalized estimator using the equation proposed by \citet[][Equation 3]{Banerjee2008}.
All computations were carried out in MATLAB R2010a, making an external call to the R package \texttt{glasso} \citep{Friedman2008}.
Table~\ref{Tab:methods} gives abbreviations for all methods and regimes investigated, which are used below and in figures.

\begin{table}
	\centering
	\caption{Clustering methods and regimes investigated, with corresponding abbreviations.}\label{Tab:methods}
	{
	\setlength{\extrarowheight}{4pt}
		\begin{tabular}{lccc}
		
		\hline
		 \multirow{2}{4cm}{\centering \textbf{Method}}       &      \multirow{2}{4cm}{\centering \textbf{Penalty term}}   &   \multirow{2}{3cm}{\centering \textbf{Tuning parameter selection} }    &   \multirow{2}{2cm}{\centering \textbf{Abbrev.}}   \\
		& & & \\
		\hline
		
		\multirow{4}{4cm}{mixture of $\ell_1$-penalized Gaussian graphical models with EM \newline (`soft' assignments)} & \multirow{2}{4cm}{\centering $p_{\lambda,\gamma}(\ThetaB)$ with $\gamma=0$: $\lambda\sum_{k=1}^K {\left\| \OmegaB_k \right\|_1}$} & \multirow{1}{3cm}{\centering Train/test} & \multirow{1}{2cm}{\centering T0} \\\cline{3-4}
		
		&  &  \multirow{1}{3cm}{\centering BIC}  & \multirow{1}{2cm}{\centering B0} \\\cline{2-4}
		&  \multirow{2}{4cm}{ \centering $p_{\lambda,\gamma}(\ThetaB)$ with $\gamma=1$: $\lambda\sum_{k=1}^K {\pi_k \left\| \OmegaB_k \right\|_1}$} & \multirow{1}{3cm}{\centering Train/test} & \multirow{1}{2cm}{\centering T1} \\\cline{3-4}
	  &  &  \multirow{1}{3cm}{\centering BIC}  & \multirow{1}{2cm}{\centering B1} \\\hline
	  
	  \multirow{3}{4cm}{$\ell_1$-penalized Gaussian graphical models \newline (`hard' assignments) \newline { \scriptsize \citet{Mukherjee2011}}} & \multirow{2}{4cm}{\centering $\lambda{\left\| \OmegaB_k \right\|_1}$, $k=1,\ldots,K$} & \multirow{1}{3cm}{\centering Train/test} & \multirow{1}{2cm}{\centering Th} \\\cline{3-4}
		
		&  &  \multirow{1}{3cm}{\centering BIC}  & \multirow{1}{2cm}{\centering Bh} \\\cline{2-4}
		
		& \multirow{1}{4cm}{\centering $\lambda_k{\left\| \OmegaB_k \right\|_1}$, $k=1,\ldots,K$} & \multirow{1}{3cm}{\centering Analytic\hyperlink{tablefootnote}{$^1$}} & \multirow{1}{2cm}{\centering Ah} \\\hline
		\multirow{1}{4cm}{ K-means} & \multirow{1}{4cm}{\centering n/a} &\multirow{1}{3cm}{\centering  n/a} & \multirow{1}{2cm}{\centering KM} \\\hline
		
		\multirow{2}{4cm}{non-penalized Gaussian mixture model with EM} & \multirow{2}{4cm}{\centering n/a} & \multirow{2}{3cm}{\centering n/a} & \multirow{2}{2cm}{\centering NP} \\
		& & & \\\hline	
		
		\end{tabular}
		\vspace{-0.2cm}
\begin{flushleft}
{\tiny \hypertarget{tablefootnote}{$^1$}following \citet{Banerjee2008} (see text for details)}
\end{flushleft}
		}
\end{table}

\subsection{Tuning parameter selection}

\begin{sidewaystable}
 \captionsetup{width=8in}
 \centering
 \caption{Simulated data; tuning parameter values selected for the methods and regimes in Table~\ref{Tab:methods}.
 Three data dimensions ($p=25,50,100$) and five per-cluster sample sizes ($n_k=15,25,50,100,200$) were considered.
 For the $\gamma=0$ regimes (T0/B0) EM update tuning parameters $\tilde{\lambda}_k$, given in (\ref{EMparam}), are also shown. 
 (Results shown are mean values over 50 simulated datasets for each $(p,n_k)$ regime;
 standard deviations given in parentheses;
 cluster-specific tuning parameter $\lambda_1$ and EM update parameter $\tilde{\lambda}_1$ correspond to the largest cluster.)}
 \label{Tab:lambda}
\scriptsize
 \begin{tabular}{cccccccccccccc}
\hline

\multirow{2}{*}{$\pB$} & \multirow{2}{*}{$\nB_{\kB}$} & \multicolumn{3}{c}{\textbf{T0}} & \multicolumn{3}{c}{\textbf{B0}} & \textbf{T1} & \textbf{B1} & \textbf{Th} & \textbf{Bh} & \multicolumn{2}{c}{\textbf{Ah}} \\
                   &                      & $\lambda$ & $\tilde{\lambda}_1$ & $\tilde{\lambda}_2$ & $\lambda$ & $\tilde{\lambda}_1$ & $\tilde{\lambda}_2$ & $\lambda$ (=$\tilde{\lambda}_k$) & $\lambda$ (=$\tilde{\lambda}_k$) & $\lambda$ & $\lambda$ & $\lambda_1$ & $\lambda_2$ \\\hline
\multirow{5}{*}{25} & 15 & 0.25 (0.07)&0.31 (0.07)&1.61 (0.73)&0.57 (0.11)&0.67 (0.12)&4.05 (0.88)&0.43 (0.08)&0.91 (0.15)&0.44 (0.11)&0.98 (0.16)&1.82 (0.43)&3.80 (1.54)\\
& 25 & 0.17 (0.04)&0.20 (0.04)&1.84 (0.82)&0.48 (0.07)&0.53 (0.08)&5.53 (0.86)&0.33 (0.05)&0.70 (0.11)&0.35 (0.10)&0.76 (0.14)&1.39 (0.23)&3.12 (1.42)\\
& 50 & 0.10 (0.01)&0.18 (0.04)&0.62 (0.94)&0.26 (0.13)&0.31 (0.10)&5.29 (3.99)&0.19 (0.03)&0.39 (0.06)&0.25 (0.03)&0.40 (0.07)&1.14 (0.43)&2.01 (1.15)\\
& 100 & 0.05 (0.00)&0.10 (0.00)&0.10 (0.00)&0.08 (0.04)&0.14 (0.05)&0.72 (2.40)&0.11 (0.02)&0.23 (0.03)&0.13 (0.03)&0.23 (0.03)&0.81 (0.13)&1.14 (0.58)\\
& 200 & 0.05 (0.00)&0.10 (0.00)&0.10 (0.00)&0.06 (0.02)&0.12 (0.04)&0.12 (0.04)&0.08 (0.02)&0.15 (0.01)&0.08 (0.02)&0.15 (0.01)&0.61 (0.08)&0.70 (0.12)\\\hline
\multirow{5}{*}{50} & 15 &0.32 (0.06)&0.38 (0.06)&2.24 (0.61)&0.66 (0.09)&0.77 (0.10)&4.65 (0.72)&0.51 (0.08)&1.03 (0.12)&0.48 (0.09)&1.12 (0.13)&2.06 (0.38)&4.81 (2.16)\\
& 25 & 0.23 (0.04)&0.26 (0.04)&2.56 (0.74)&0.53 (0.06)&0.58 (0.06)&6.06 (0.76)&0.38 (0.08)&0.81 (0.10)&0.37 (0.09)&0.84 (0.11)&1.63 (0.31)&4.39 (1.42)\\
& 50 & 0.14 (0.03)&0.16 (0.02)&2.60 (1.53)&0.36 (0.03)&0.38 (0.03)&8.21 (1.16)&0.29 (0.02)&0.53 (0.08)&0.34 (0.06)&0.53 (0.06)&1.18 (0.18)&4.32 (1.78)\\
& 100 & 0.07 (0.02)&0.10 (0.02)&1.32 (1.89)&0.25 (0.03)&0.25 (0.02)&10.50 (2.26)&0.15 (0.00)&0.30 (0.02)&0.19 (0.02)&0.30 (0.02)&0.90 (0.35)&3.06 (1.70)\\
& 200 & 0.05 (0.00)&0.10 (0.00)&0.10 (0.00)&0.05 (0.01)&0.10 (0.01)&0.10 (0.01)&0.10 (0.00)&0.20 (0.00)&0.10 (0.00)&0.20 (0.00)&0.67 (0.36)&1.84 (1.38)\\\hline
\multirow{5}{*}{100} & 15 & 0.39 (0.04)&0.45 (0.05)&2.69 (0.38)&0.72 (0.06)&0.84 (0.07)&5.04 (0.58)&0.55 (0.06)&1.20 (0.12)&0.55 (0.06)&1.23 (0.11)&2.70 (1.04)&5.30 (1.91)\\
& 25 & 0.28 (0.03)&0.31 (0.03)&3.07 (0.51)&0.57 (0.05)&0.63 (0.06)&6.54 (0.83)&0.42 (0.06)&0.96 (0.12)&0.41 (0.04)&0.99 (0.09)&2.22 (0.68)&6.50 (3.07)\\
& 50 & 0.19 (0.02)&0.20 (0.02)&4.28 (0.74)&0.40 (0.04)&0.42 (0.04)&8.46 (1.56)&0.37 (0.05)&0.59 (0.04)&0.28 (0.04)&0.61 (0.05)&1.47 (0.35)&5.11 (1.60)\\
& 100 & 0.10 (0.01)&0.11 (0.01)&4.04 (1.07)&0.27 (0.02)&0.27 (0.02)&11.30 (1.86)&0.21 (0.02)&0.36 (0.02)&0.27 (0.02)&0.37 (0.03)&1.04 (0.24)&5.75 (2.75)\\
& 200 & 0.05 (0.00)&0.10 (0.00)&0.10 (0.00)&0.20 (0.01)&0.20 (0.01)&15.48 (2.60)&0.11 (0.02)&0.25 (0.01)&0.15 (0.00)&0.25 (0.01)&0.78 (0.20)&4.58 (2.40)\\\hline
\end{tabular}
\end{sidewaystable}

Table~\ref{Tab:lambda} shows average tuning parameter values selected by each regime.
A grid search was used over values between 0.05 and 1.5, with increments of 0.05.
Since, for $\gamma=0$, the EM update tuning parameters $\tilde{\lambda}_k$ in (\ref{ICk}) differ from $\lambda$, we also show $\tilde{\lambda}_k$ for these regimes. 
Using BIC to set the tuning parameter results in higher values than with train/test and, as expected, $\lambda$ values increase with $p$ and decrease with $n_k$.

\subsection{Cluster assignment}  

 	 \begin{sidewaysfigure}[tbp]
	\centering
	\subfloat[p=25]{
		\includegraphics[width=\textwidth]{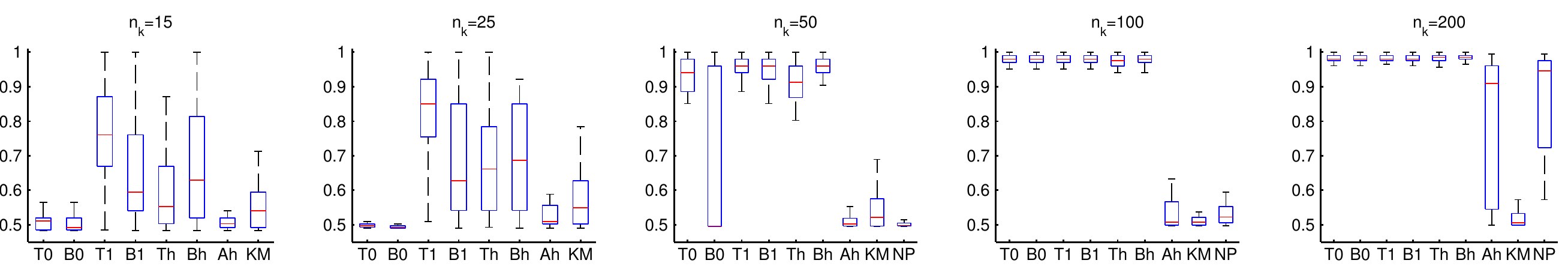}}\\
	\subfloat[p=50]{
		\includegraphics[width=\textwidth]{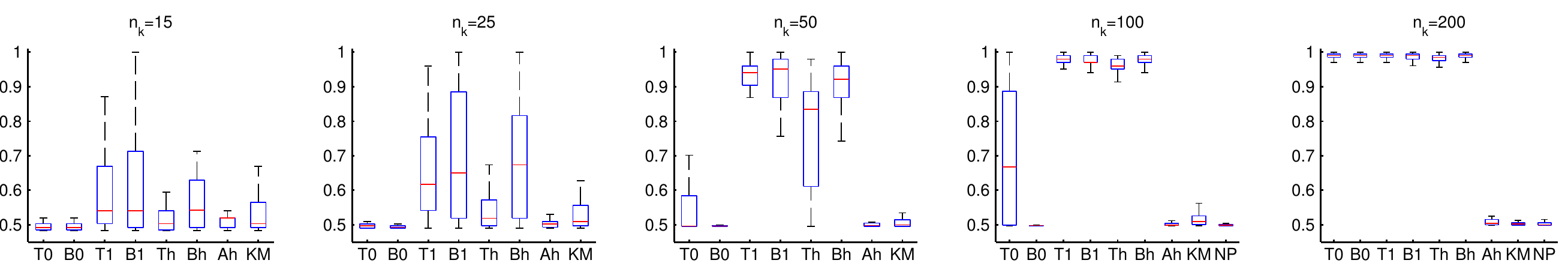}}\\
	\subfloat[p=100]{
		\includegraphics[width=\textwidth]{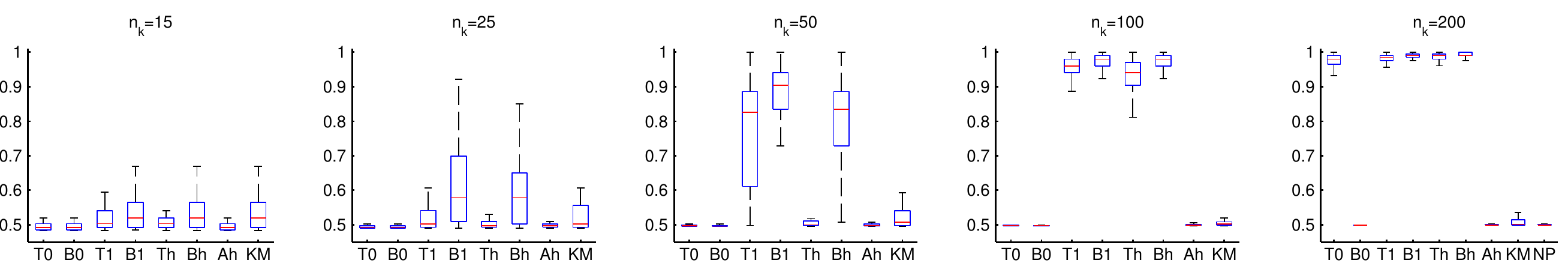}}
	\caption{Simulated data; cluster assignment results.
	Boxplots over the Rand index, a measure of similarity between inferred and true cluster labels (higher scores indicate better agreement, with a score of unity indicating perfect agreement), are shown for the methods and regimes in Table~\ref{Tab:methods} at varying data dimensions $p$ and per-cluster sample sizes $n_k$. (Results shown are over 50 simulated datasets for each $(p,n_k)$ regime (see text for details);  abbreviations for methods are summarized in Table~\ref{Tab:methods}; the non-penalized approach (NP) could not be used for $n_k\leq p$ due to small sample sizes resulting in invalid covariance estimates.)}
	\label{fig:RI}
\end{sidewaysfigure}

   Figure~\ref{fig:RI} shows Rand indices (with respect to the true cluster assignments) obtained from clustering the simulated data.
The Rand index is a measure of similarity between cluster assignments, taking values between 0 and 1 (0 indicates complete disagreement and 1 complete agreement).	
Box plots are shown over 50 simulated datasets for each $(p,n_k)$ regime.
	 The $\ell_1$-penalized mixture model regimes with $\gamma=1$ in the penalty term (T1/B1) consistently provide the best clustering results.
At the largest sample sizes both train/test (T1) and BIC (B1) offer good clustering performance, with high Rand indices reported.
However, for smaller sample sizes, train/test outperforms BIC at the lowest data dimensionality ($p=25$), while the converse is true at higher dimensions ($p=50,100$).
   The non-mixture $\ell_1$-penalized method (Th/Bh) also performs well, but the corresponding mixture model approaches with $\gamma=1$ (T1/B1) are, for the most part, more effective at smaller sample sizes (see e.g. $n_k=50$, $p=50,100$).
This difference in performance is likely due to a combination of differences in tuning parameter (Supplementary Table~S1) and less accurate parameter estimation for the non-mixture approaches because they do not take uncertainty of  assignment into account.
	 Interestingly, the mixture model with conventional penalty term ($\gamma=0$; T0/B0) shows poor performance relative to $\gamma=1$ except at larger sample sizes, with  consistently  poor clustering accuracy for $n_k\leq p$.
Similar performance is observed for the non-mixture method with analytic tuning parameter selection (Ah).
The poor performance of these three regimes appears to be related to the fact that they all use cluster-specific tuning parameters ($\lambda_k$ for Ah and $\tilde{\lambda}_k$ within EM for T0/B0), resulting in considerable differences in cluster-level penalties (see Supplementary Table~S1).
We comment further on this finding in Discussion below.	 
   Due to its inability to capture the cluster-specific covariance (network) structure, K-means does not perform well, even at the largest sample size.
   Conventional non-penalized mixture models did not yield valid covariance estimates for sample sizes  $n_k\leq p$, and for $n_k>p$ we only observe gains relative to K-means in the large sample $p=25$, $n_k=200$ case.

\subsection{Estimation of  graphical model structure}  	

\begin{figure}[tbp]
	\centering
	\subfloat[p=25]{
		\includegraphics[width=0.8\textwidth]{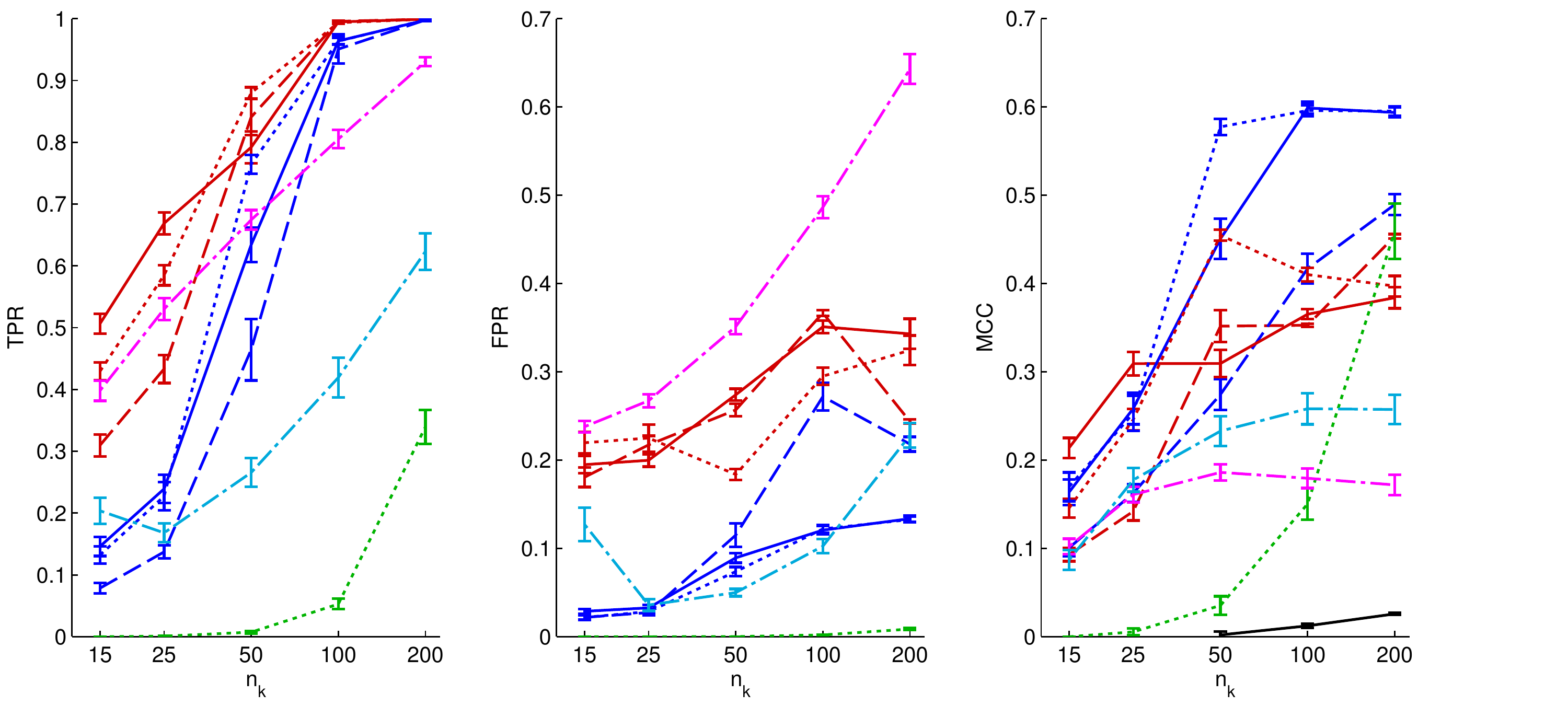}}\\
	\subfloat[p=50]{
		\includegraphics[width=0.8\textwidth]{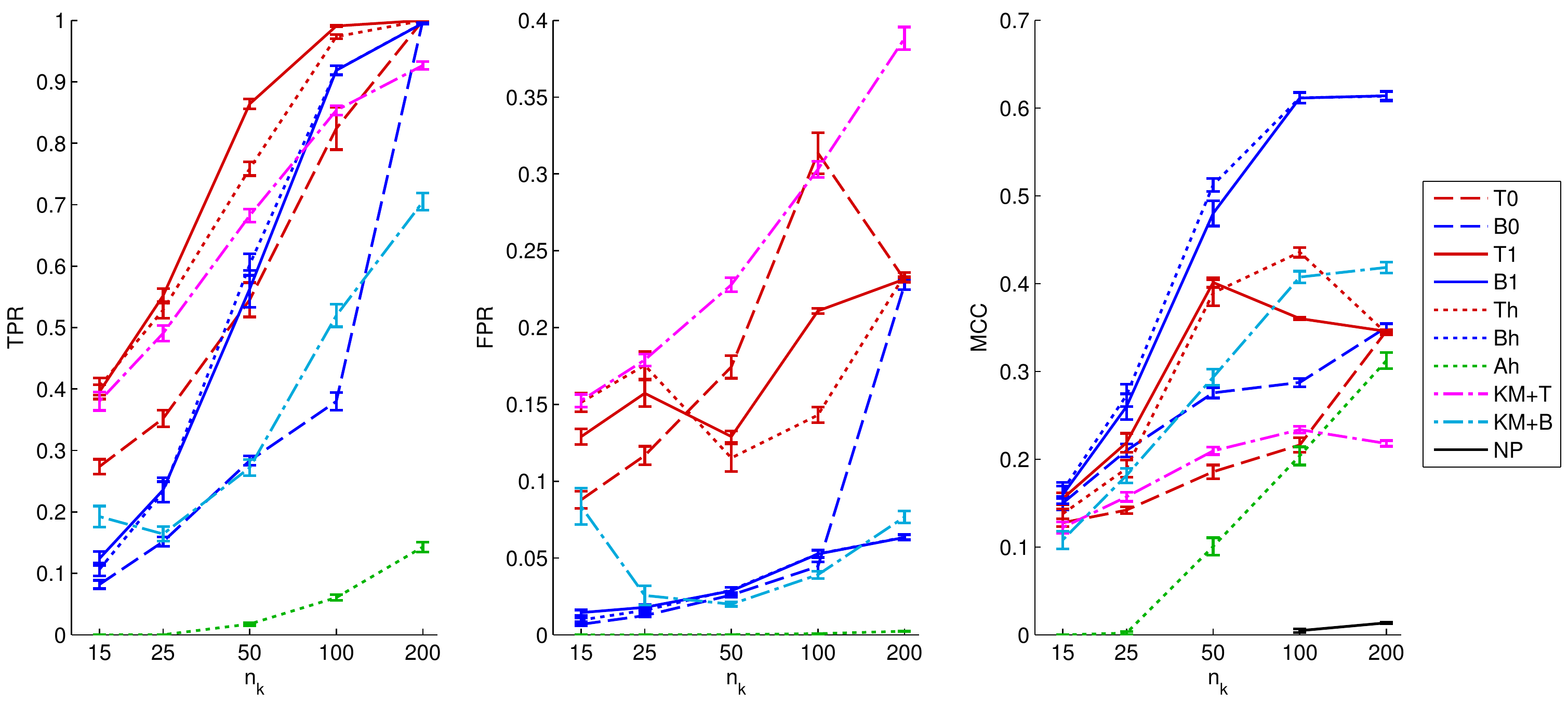}}\\
	\subfloat[p=100]{
		\includegraphics[width=0.8\textwidth]{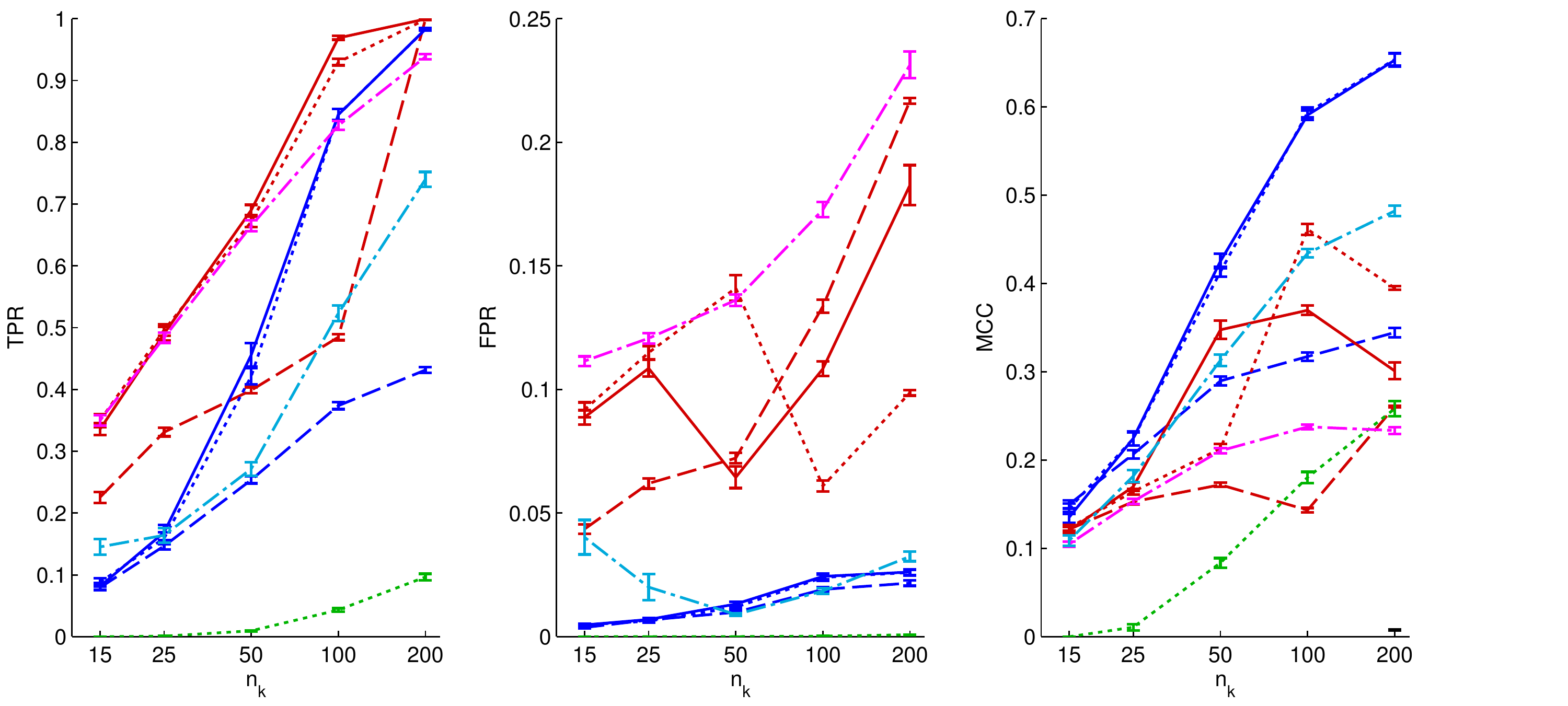}}
	\caption{Simulated data; estimation of graphical model  structure.
	True Positive Rate (TPR), False Positive Rate (FPR) and Matthews Correlation Coefficient (MCC) are shown as a function of per-cluster sample size $n_k$ for the methods and regimes in Table~\ref{Tab:methods} and at data dimensions $p=25,50,100$.
	MCC is a balanced measure for classification performance, taking values between -1 and 1 with higher values indicating better agreement between true and inferred networks (see text for details).
	K-means clustering was followed by $\ell_1$-penalized estimation of Gaussian graphical model structure with penalty parameter set by train/test (`KM+T') or BIC (`KM+B').
	(Mean values shown over 50 simulated datasets for each $(p,n_k)$ regime, error bars show standard errors; non-penalized approach (NP) is only shown for MCC and could not be used for $n_k\leq p$ due to small sample sizes resulting in invalid covariance estimates.)}
	\label{fig:GGMest}
\end{figure}

Figure~\ref{fig:GGMest} shows results for estimation of cluster-specific network structures for the
methods and regimes in Table~\ref{Tab:methods}.
For K-means, clustering is followed by an application, to each inferred cluster, of $\ell_1$-penalized precision matrix estimation (see (\ref{pLL1})) with tuning parameter set by either BIC or train/test.

Ability to reconstruct cluster-specific networks  is assessed by calculating the true positive rate (TPR),  false positive rate (FPR) and Matthews Correlation Coefficient (MCC),
\begin{gather}
TPR=\frac{TP}{TP+FN}, \ \ \ \ \ \ \ \  \ \ \ FPR=\frac{FP}{FP+TN} \\[1em]
MCC = \frac{TP\times TN - FP\times FN}{\sqrt{(TP+FP)(TP+FN)(TN+FP)(TN+FN)}}\label{MCC}
\end{gather}
where $TP$, $TN$, $FP$ and $FN$ denote the number of true positives, true negatives, false positives and false negatives (with respect to edges) respectively.
MCC summarizes these four quantities into one score and is regarded as a balanced measure; it takes values between -1 and 1, with higher values indicating better performance (see e.g. \citet{Baldi2000} for further details).
Since the convergence threshold in the \texttt{glasso} algorithm is $10^{-4}$, we take entries  $\hat{\omega}_{ij}$ in  estimated precision matrices to be non-zero if $\left|\hat{\omega}_{ij}\right|>10^{-3}$.
Since cluster assignments can only be identified up to permutation, in all cases labels were permuted to maximize agreement with true cluster assignments before calculating these quantities.

Figure~\ref{fig:GGMest} shows MCC plotted against per-cluster sample size $n_k$
and Supplementary Figure~S1 shows corresponding plots for TPR and FPR.
Due to selection of smaller tuning parameter values, BIC discovers fewer non-zeroes in the precision matrices than train/test, resulting in both fewer true positives and false positives.
Under MCC, BIC, with either the $\gamma=1$ mixture model (B1) or the non-mixture approach (Bh), leads to  the best network reconstruction (except at small sample sizes with $p=25$) 
and  outperforms all other regimes at larger sample sizes.

In general, train/test is not competitive relative to BIC; at larger sample sizes the best train/test regimes (T1/Th) are only comparable with the worst performing BIC regimes (B0/KM+B).
We note that the non-penalized mixture approach (NP), with sample size sufficiently large to provide valid covariance estimates, does not yield sparse precision matrices (MCC scores are approximately zero).

\subsection{Precision matrix estimation}

\begin{table}[tbp]
 \centering 
 \setlength{\tabcolsep}{4pt}
 \caption{Simulated data; precision matrix estimation.
 Elementwise $\ell_1$ matrix norm 
 shown for the methods and regimes in Table~\ref{Tab:methods} at varying data dimensions $p$ and per-cluster sample sizes $n_k$ (smaller values indicate better agreement between true and inferred precision matrices).
 K-means clustering was followed by $\ell_1$-penalized estimation of Gaussian graphical model structure with penalty parameter set by train/test (`KM+T') or BIC (`KM+B').
 For each $(p,n_k)$ combination, the regime with lowest mean norm is highlighted in bold.
(Mean matrix norm over 50 simulated datasets per $(p,n_k)$ regime, standard deviations in parentheses;
non-penalized, full covariance Gaussian mixture model (NP) could not be used for small sample sizes $n_k\leq p$ due to invalid covariance estimates.)
 }
 \label{tab:l1norm}
 \small
 \begin{tabular}{cccccccccccc}

\hline

\multirow{1}{*}{\centering  $\pB$} &\multirow{1}{*}{\centering  $\nB_{\kB}$} &\textbf{T0} & \textbf{B0} & \textbf{T1} & \textbf{B1} & \textbf{Th} & \textbf{Bh} & \textbf{Ah} & \textbf{KM+T} & \textbf{KM+B} & \textbf{NP} \\ \hline

\multirow{11}{*}{\centering  25} & \multirow{2}{*}{\centering 15} & 64.00 &65.73 & \textbf{56.88 }&58.92 &60.77 &59.71 &69.21 &62.05 &158.01 &\multirow{2}{*}{-}\\
& & (4.00) & (2.52)&  (4.87)&(3.12)& (5.59)& (3.07)& (2.58)& (4.35)& (136.36) &  \\[0.6mm]

& \multirow{2}{*}{\centering  25} & 64.41 &66.39 &\textbf{52.36 }&56.64 &57.79 &57.35 &67.31 &60.02 &62.92 &\multirow{2}{*}{-}\\
& & (4.38)& (2.53)& (4.47)&(3.61)&(6.48)& (4.01)& (3.63)& (4.35)& (31.35) & \\ [0.6mm]

& \multirow{2}{*}{\centering  50} & 47.56 &57.83 &47.76 &49.89 &\textbf{44.06 }&47.51 &64.50 &60.15 &56.20 &814.79 \\
& & (9.87)&(11.67)& (6.83)&(4.01)& (2.45)&(3.36)& (3.28)& (6.37)& (4.22)&(282.61) \\ [0.6mm]

& \multirow{2}{*}{\centering  100} & 35.53&37.73&35.43&38.22&\textbf{35.15}&38.04&61.41&62.78&54.86&275.75 \\
& & (1.86)& (7.06)& (1.88)& (2.42)& (1.95)& (2.35)& (3.50)& (8.74)& (5.87)& (29.26)\\[0.6mm] 

& \multirow{2}{*}{\centering  200} & \textbf{26.56}&27.81&27.86&29.73&27.57&29.64&56.68&64.78&53.09&112.64\\
& & (1.48)& (3.22)& (2.56)& (1.59)& (2.52)& (1.66)& (3.18)& (11.49)& (7.38)& (21.30)\\ [0.6mm]
\hline

\multirow{11}{*}{\centering  50} & \multirow{2}{*}{\centering  15} & 126.47&129.52&119.20&\textbf{117.25}&123.93&119.21&137.91&124.41&450.23&\multirow{2}{*}{-}\\
& & (5.57)& (5.09)& (6.54)& (5.25)& (9.17)& (5.60)& (5.81)& (6.25)& (377.88)&  \\[0.6mm]

& \multirow{2}{*}{\centering  25} & 129.08&129.75&118.52&\textbf{113.23}&123.89&114.08&136.01&124.50&159.21&\multirow{2}{*}{-}\\
& & (4.89)& (4.46)& (11.53)& (5.46)& (11.13)& (5.54)& (5.25)& (6.53)& (172.28)&  \\[0.6mm]

& \multirow{2}{*}{\centering 50} & 128.38&129.29&\textbf{94.08}&103.06&102.96&102.88&134.11&123.51&112.27&\multirow{2}{*}{-}\\
& & (7.68)& (3.57)& (4.02)& (6.16)& (10.57)& (5.00)& (5.76)& (6.45)& (4.56)& \\[0.6mm]

& \multirow{2}{*}{\centering 100} &120.24&125.51&78.95&82.77&\textbf{77.24}&82.90&127.33&121.14&104.06&1366.07\\
& &  (14.90)& (7.66)& (3.26)& (3.12)& (3.56)& (3.51)& (7.37)& (7.77)& (4.46)& (96.11)\\[0.6mm]

& \multirow{2}{*}{\centering 200} & 63.69&63.79&\textbf{63.67}&67.31&64.10&67.15&119.91&122.92&98.90&587.84\\
& &  (2.70)& (2.95)& (2.69)& (2.44)& (2.73)& (2.46)& (8.74)& (10.50)& (5.39)& (23.48)\\[0.6mm]
\hline

\multirow{11}{*}{\centering 100} & \multirow{2}{*}{\centering 15} & 250.78&256.29&247.53&\textbf{236.25}&249.46&237.79&277.45&262.41&799.86&\multirow{2}{*}{-}\\
& &  (8.60)& (8.09)& (13.87)& (10.02)& (14.67)& (9.50)& (9.77)& (10.77)& (794.03)&  \\[0.6mm]

& \multirow{2}{*}{\centering 25} & 256.42&256.42&249.49&\textbf{229.69}&253.52&231.93&277.61&259.05&427.52&\multirow{2}{*}{-}\\
& &  (7.57)& (7.66)& (14.25)& (9.76)& (12.55)& (8.59)& (9.50)& (11.26)& (534.15)&  \\[0.6mm]

& \multirow{2}{*}{\centering 50} & 254.02&249.76&\textbf{203.44}&205.45&248.40&208.04&267.44&244.39&216.46&\multirow{2}{*}{-}\\
& &  (8.77)& (6.96)& (16.47)& (6.45)& (23.35)& (7.93)& (10.53)& (10.51)& (9.12)&  \\[0.6mm]

& \multirow{2}{*}{\centering 100} & 269.91&246.16&172.36&176.36&\textbf{167.65}&176.92&263.57&238.25&203.45&\multirow{2}{*}{-}\\
& &  (9.40)& (8.07)& (6.88)& (6.08)& (5.26)& (6.43)& (12.30)& (12.31)& (9.27)&  \\[0.6mm]

& \multirow{2}{*}{\centering 200} & 171.59&239.45&158.82&144.53&\textbf{134.00}&144.68&252.23&235.31&183.98&3468.41\\
& & (11.72)& (6.33)& (17.70)& (4.69)& (3.76)& (4.31)& (15.93)& (19.05)& (8.16)& (170.53)\\[0.6mm]
 \hline

\end{tabular}
\end{table}

We also assessed ability to accurately estimate underlying cluster-specific precision matrices (i.e. values of the matrix elements rather than only locations of non-zeros).
Accuracy is assessed using the elementwise $\ell_1$ norm, $\sum_{k=1}^K \left\|\hat{\OmegaB}_k - \OmegaB_k \right\|_1$, with inferred clusters  matched to true clusters as described above.
Results are shown in  Table~\ref{tab:l1norm}.
In contrast to clustering and Gaussian graphical model estimation, where BIC regimes B1/Bh mainly provide the best performance, the train/test methods T1/Th are mostly similar or better than B1/Bh for precision matrix estimation (the exception being small $n_k$,  higher $p$ settings). 
Due to poor clustering performance, the mixture model approach with $\gamma=0$ does not perform well unless $n_k$ is sufficiently large.
Neither K-means clustering (followed by $\ell_1$-penalized precision matrix estimation), the penalized non-mixture approach with analytic tuning parameter selection (Ah), nor the non-penalized approach (NP) perform well, even at the largest sample size.

\subsection{Approximate tuning parameter selection}

  	 \begin{figure}[tbp]
	\centering
	\subfloat[Tuning parameter $\lambda$]{
		\includegraphics[width=\textwidth]{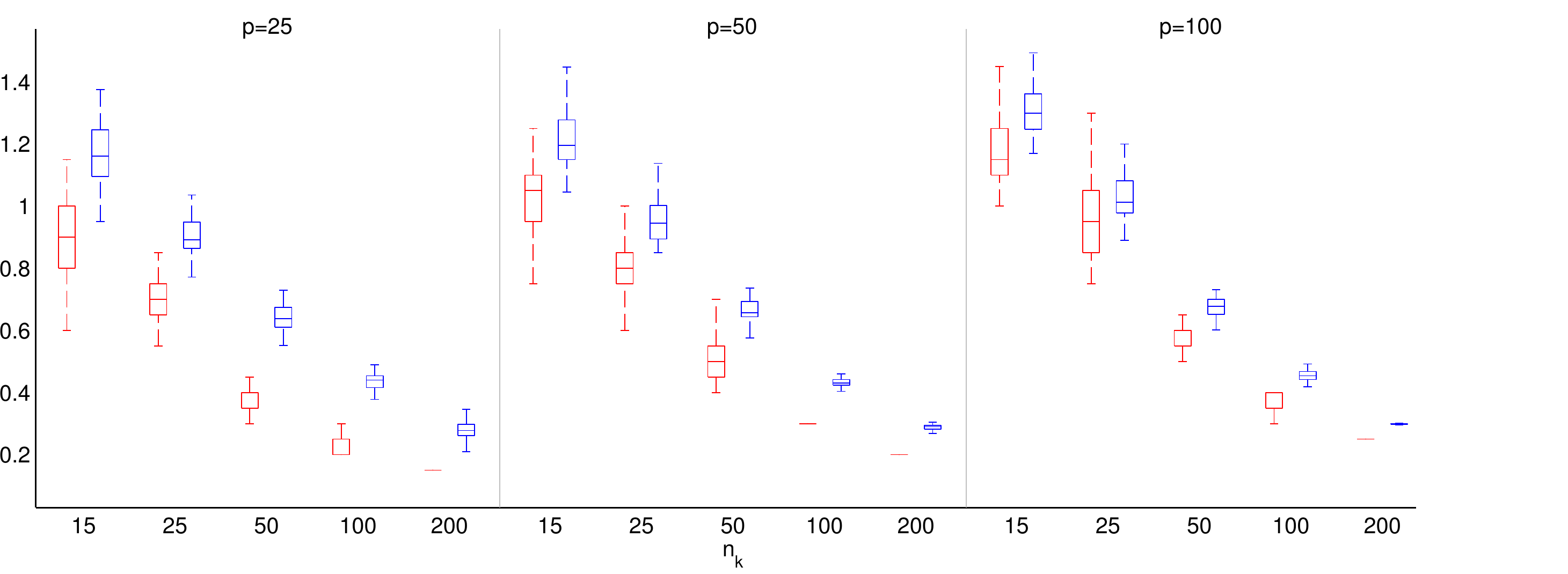}}\\
	\subfloat[Rand index]{
		\includegraphics[width=\textwidth]{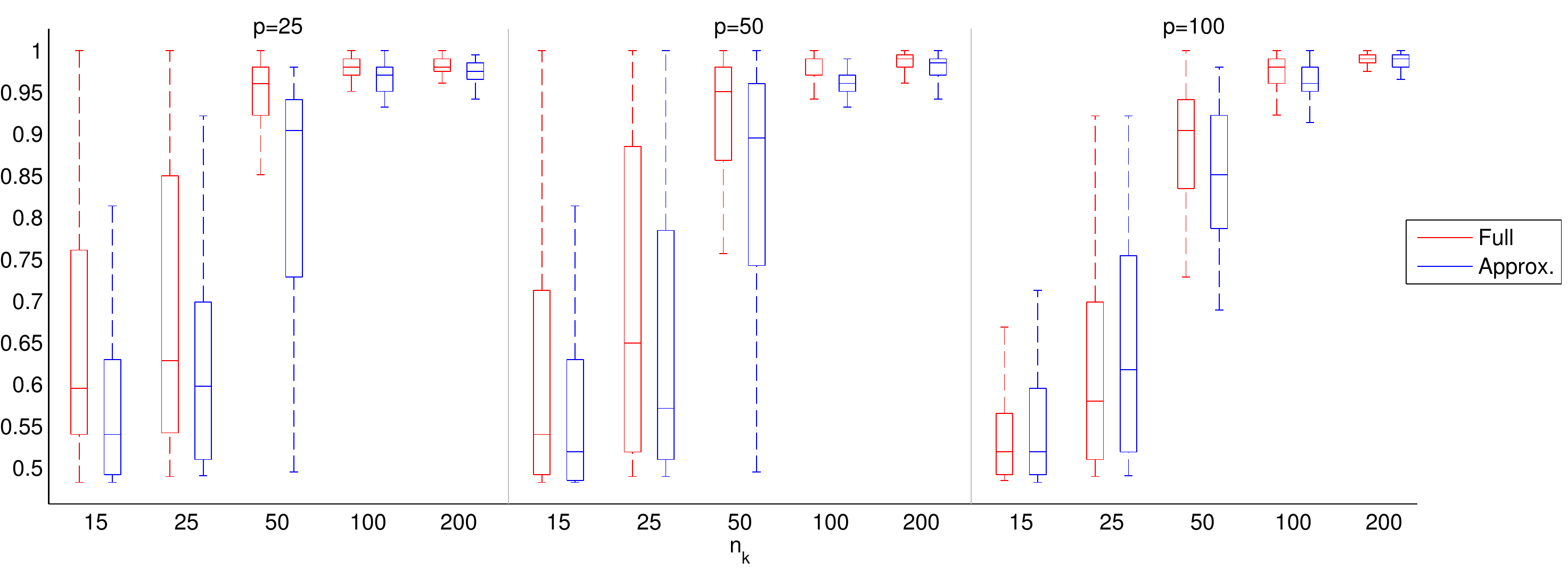}}\\
	\subfloat[Computation time (seconds)]{
		\includegraphics[width=\textwidth]{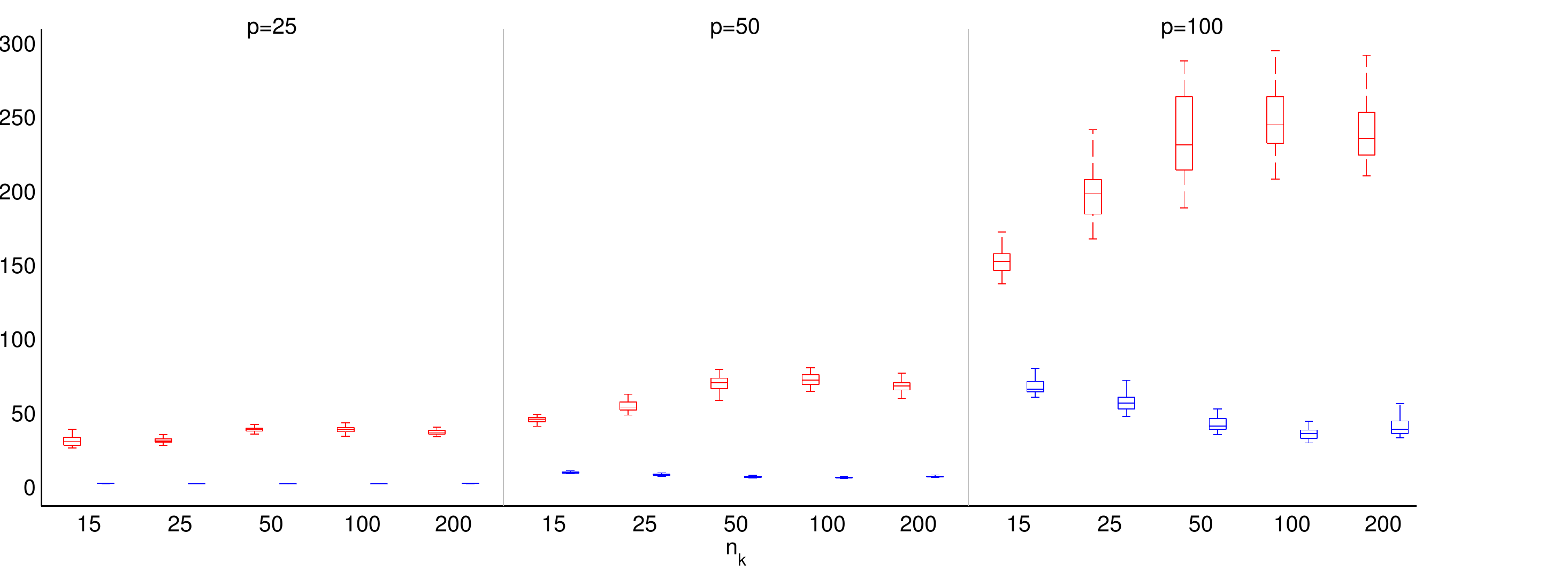}}
	\caption{{ \bf Simulated data; heuristic approach for tuning parameter selection.}
	(a) Boxplots over the tuning parameters selected by the heuristic method (see text for details) under regime B1 (mixture model with $\gamma=1$ and BIC) are shown (blue boxes), together with the corresponding values obtained with the full, non-approximate approach (red boxes).
 (b) Resulting Rand indices and (c) computational time required to set the parameter are also shown.
 (All results are over 50 simulated datasets for each $(p,n_k)$ regime).}
	\label{fig:approx}
\end{figure}

We applied the heuristic method for setting the tuning parameter, described in Methods above,
to the overall best-performing mixture model approach (regime B1; $\gamma=1$, BIC).
Figure~\ref{fig:approx} compares average $\lambda$ values obtained using the heuristic method with those resulting from the full approach;
we also show average Rand indices and computational timings.
The $\lambda$ values obtained via the heuristic scheme are well-behaved in the sense that they 
increase with $p$ and decrease for larger $n_k$.
We observe some bias  relative to the full approach as the values obtained from the heuristic method are consistently higher.
However, Rand indices remain in reasonable agreement and the heuristic offers some substantial computational gains; e.g. for $p=25$ we see reduction of about 90\% in computation time. This suggests that the heuristic approach could be useful for fast, exploratory analyses.

\section{Discussion}
We presented a study of model-based clustering with mixtures of $\ell_1$-penalized Gaussian graphical models.
Penalization has emerged as a key approach in high-dimensional statistics.
However, choice of penalization set-up and the setting of tuning parameters can be non-trivial.
We found that performance is dependent on choice of penalty term and method for setting the tuning parameter.
Along with the standard $\ell_1$ penalty ($\gamma=0$ in (\ref{penterm})) we considered an alternative penalty term, following recent work in penalized finite mixture of regression models \citep{Khalili2007, Stadler2010}, that is dependent on the mixing proportions $\pi_k$ ($\gamma=1$ in (\ref{penterm})).

 From our simulation study and application to breast cancer data, we draw some broad conclusions and recommendations, as follows.
The combination of the $\gamma=1$ penalty term (incorporating mixing proportions), together with the BIC criterion for selecting the tuning parameter (regime B1), appears to provide the most accurate clustering and estimation of  graphical model structure.
The only exception  is in settings where both dimensionality and sample size are small; here, the smaller tuning parameter values selected by train/test (or cross-validation) provide superior results (regimes T1/CV1).
For estimation of the precision matrix itself (as opposed to estimation of sparsity structure only), we again recommend the penalty term with $\gamma=1$ and find that the less sparse estimates provided by train/test (or cross-validation) provide slight gains over BIC, except where dimensionality is  large relative to sample size.

The deleterious effect of the standard $\ell_1$ penalty term ($\gamma=0$), at all but the largest sample sizes, is intriguing.
As described above, this is due to the fact that the standard penalty term leads to cluster-specific penalties in the EM update for the precision matrices.
(Indeed, we observed similar results when setting cluster-specific penalties analytically in a non-mixture model setting).
These cluster-specific penalties are inversely proportional to the mixing proportions $\pi_k$:
in itself this behavior seems intuitively appealing since clusters with small effective sample sizes are then more heavily regularized.
However, we observed that a substantially higher penalty is applied to one cluster over the other, indicating that samples were mostly being assigned to the same cluster.
This is likely due to the `unpopular' cluster having a poor precision matrix estimate due to a large penalty.
We note that this behavior is not due to (lack of) EM convergence; the penalized likelihood scores from these incorrect clusterings were higher than those obtained using the true cluster labels.

The related non-mixture model approach proposed by \citet{Mukherjee2011} also performed well in our studies, but clustering results (both from simulated and real data) indicate that a mixture model with EM (and $\gamma=1$ in the penalty term) offers more robust results.

Although the approaches we recommend performed well in the examples we considered, sensitivity to penalty formulation and the setting of tuning parameters remain a concern for penalized mixtures. Further work will be needed to better understand how such approaches behave in other settings and in higher dimensions.  Cluster-specific scaling could also pose difficulties for penalization, as discussed recently in the context of hidden Markov models in \citet{Stadler2012}, who propose penalisation using the inverse correlation matrix as a potential solution.
The approach proposed here could be adapted to use inverse correlation in place of inverse covariance. 

We note that while for simplicity and tractability we focused on the $K=2$ clusters case, the methods  we discuss are immediately applicable to the general $K$-cluster case.
Moreover, since the approach we propose is model-based, established approaches for  model selection in clustering, including information criteria, train/test and cross-validation,
 can be readily employed to select or explore $K$.

Our results  demonstrate the necessity of some form of regularization to enable the use of Gaussian graphical models for clustering in settings of moderate-to-high dimensionality; indeed, we see clear benefits of penalization already in the $p=25$ case.
The $\ell_1$-penalty is an attractive choice since it encourages sparsity at the level of graphical model structure, and estimation with the graphical lasso algorithm \citep{Friedman2008} is particularly efficient, which is important in the clustering setting, where multiple iterations are required.
Alternatives 
include shrinkage estimators \citep{Schafer2005} and Bayesian approaches \citep{Dobra2004,Jones2005}.
However, it has been shown that the $\ell_1$-penalized precision matrix estimator (\ref{pLL1}) is biased \citep{Lam2009}.
Alternative penalties have been proposed in a regression setting to ameliorate this issue; the non-concave SCAD penalty \citep{Fan2001} and adaptive $\ell_1$ penalty \citep{Zou2006}, and have recently been applied to sparse precision matrix estimation \citep{Fan2009}.
These penalties are generally  computationally more intensive, but it remains an open question whether they improve clustering accuracy relative to the $\ell_1$ penalty considered here.

Graphical models based on direct acyclic graphs (DAGs) are frequently used for network inference, especially in biological settings where directionality may be meaningful \citep[for example,][]{Friedman2000, Husmeier2003, Perrin2003, Mukherjee2008, Ellis2008,Hill2012c}.
A natural extension to the ideas discussed here would be to develop a clustering approach based on DAGs rather than  undirected models.

There are several recent and attractive extensions to graphical Gaussian model estimation that could be exploited to improve and extend the methods we discuss.
For example, the time-varying Gaussian graphical model approach of \citet{Zhou2010} could be employed, or prior knowledge of network structure could be taken into account \citep{Anjum2009}; such information is abundantly available in biological settings.
The joint estimation method for Gaussian graphical models proposed by \citet{Guo2011} explicitly models partial agreement between network structures corresponding to \emph{a priori} known clusters.
Such partial agreement could be incorporated in the current setting where clusters are not known \emph{a priori}.\\

\noindent {\bf Acknowledgement:} We thank P. T. Spellman, N. St\"{a}dler and N. Meinshausen for  discussions. Supported by EPSRC EP/E501311/1 and NCI U54 CA 112970 and the
Cancer Systems Biology Center grant from the Netherlands Organisation for Scientific Research.

\bibliographystyle{natbib}
\bibliography{masterrefs}

\end{document}